\documentclass{article}

\usepackage{amsmath}
\usepackage{graphicx}

\usepackage[sglblindworkshop, final]{neurips_2025}

\usepackage[utf8]{inputenc} 
\usepackage[T1]{fontenc}    
\usepackage{hyperref}       
\usepackage{url}            
\usepackage{booktabs}       
\usepackage{amsfonts}       
\usepackage{nicefrac}       
\usepackage{microtype}      
\usepackage{xcolor}         

\title{Origin-Conditional Trajectory Encoding: Measuring Urban Configurational Asymmetries through Neural Decomposition}
\workshoptitle{UrbanAI: Harnessing Artificial Intelligence for Smart Cities}

\author{%
  Stephen Law \thanks{Corresponding author}\\
  University College London\\
  London, Uk \\
  \texttt{stephen.law@ucl.ac.uk} \\
  \And
  Tao Yang \\
  Tsinghua University\\
  Beijing, China \\
  \texttt{yangtao128@tsinghua.edu.cn} \\
    \And
  Nanjiang Chen \\
  University of Sydney\\
  Sydney, Australia \\
  \texttt{nanjiang.chen@sydney.edu.au} \\
      \And
  Xuhui Lin \\
  University College London\\
  London, UK \\
  \texttt{xuhui.lin.16@ucl.ac.uk} \\
}

\begin{document}

\maketitle

\begin{abstract}
Urban analytics increasingly relies on AI-driven trajectory analysis, yet current approaches suffer from methodological fragmentation: trajectory learning captures movement patterns but ignores spatial context, while spatial embedding methods encode street networks but miss temporal dynamics. Three gaps persist:  (1) lack of joint training that integrates spatial and temporal representations, (2) origin-agnostic treatment that ignores directional asymmetries in navigation ($A \to B \ne B \to A$), and (3) over-reliance on auxiliary data (POIs, imagery) rather than fundamental geometric properties of urban space. We introduce a conditional trajectory encoder that jointly learns spatial and movement representations while preserving origin-dependent asymmetries using geometric features. This framework decomposes urban navigation into shared cognitive patterns and origin-specific spatial narratives, enabling quantitative measurement of cognitive asymmetries across starting locations. Our bidirectional LSTM processes visibility ratio and curvature features conditioned on learnable origin embeddings, decomposing representations into shared urban patterns and origin-specific signatures through contrastive learning. Results from six synthetic cities and real-world validation on Beijing's Xicheng District demonstrate that urban morphology creates systematic cognitive inequalities. This provides urban planners quantitative tools for assessing experiential equity, offers architects insights into layout decisions' cognitive impacts, and enables origin-aware analytics for navigation systems.
\end{abstract}

\section{Introduction}
Urban trajectory analysis has emerged as a cornerstone of computational urban studies, enabling researchers to decode mobility patterns, predict movement flows, and optimize transportation systems \cite{zheng2014urban, calabrese2013understanding}. The proliferation of GPS tracking devices and mobile sensing technologies has generated unprecedented volumes of human movement data, spurring the development of sophisticated machine learning methods for understanding urban dynamics \cite{song2010limits, gonzalez2008understanding}. However, despite these advances, current approaches suffer from a fundamental methodological fragmentation that limits their ability to capture the full complexity of urban spatial experience.
Contemporary urban AI operates through two largely disconnected paradigms. Trajectory representation learning methods such as T2Vec \cite{li2018deep, feng2018deepmove} excel at capturing sequential movement patterns and predicting destination choices, but treat urban space as a neutral substrate where spatial context plays no meaningful role in movement interpretation. Conversely, spatial embedding approaches including Node2Vec \cite{grover2016node2vec} and urban graph neural networks \cite{fan2024neural} effectively encode street network topology and spatial relationships, but ignore the temporal dynamics of movement and the sequential nature of navigation experiences. This methodological divide creates a critical blind spot: neither approach captures how movement and place co-evolve during navigation, missing the integrated spatiotemporal narratives that characterize human urban experience.
This fragmentation manifests in three fundamental limitations that constrain current urban analytics capabilities. First, while trajectory methods learn movement semantics and spatial methods capture network topology, no existing approach jointly optimizes both spatial and temporal representations. This separation fails to model how residents experience cities as integrated spatiotemporal environments where each navigation step is conditioned by both trajectory history and local spatial context. Joint training should produce more expressive embeddings that capture these unified urban experiences, yet remains unexplored in urban AI applications \cite{wang2021survey,jiang2023self}.
Second, current trajectory analysis treats paths as origin-independent sequences, assuming that movement patterns are equivalent regardless of starting location. This assumption conflicts with established findings in spatial cognition research demonstrating that navigation is fundamentally perspective-dependent \cite{okeefe1978hippocampus,epstein2017cognitive}. The same street traversed from different starting points creates distinct cognitive demands, landmark sequences, and wayfinding strategies \cite{lynch1960image,montello1998new}. Traditional methods would treat journeys $A \to B$ and $ \to A$ as equivalent movement patterns, failing to capture the asymmetric cognitive loads they generate. This origin-agnostic treatment obscures crucial dimensions of how cities are actually experienced by their inhabitants. Third, most urban AI systems enrich trajectory analysis with auxiliary information such as points of interest (POIs), street-view imagery, or demographic data \cite{yao2018representing,liu2020trajgat}. While these approaches demonstrate improved performance, they obscure the fundamental role of 3D spatial geometry in navigation cognition. Drawing from space syntax theory \cite{hillier1996space}, which emphasizes how built environment configuration shapes movement patterns, we argue that visibility relationships and spatial form changes—derived purely from geometric properties—capture essential cognitive elements of urban navigation. This geometric focus reveals how physical urban form directly influences navigation experience, independent of semantic or functional overlays.

We introduce a conditional trajectory encoding framework that addresses these limitations through unified neural architecture design. Our approach centers on the hypothesis that urban cognition exhibits systematic asymmetries that can be computationally measured and compared: residents from different origins develop distinct mental models of the same urban space, and these differences reflect meaningful variations in cognitive load and navigation confidence. The key innovation lies in decomposing trajectory representations into shared components that capture universal urban movement patterns and origin-specific components that encode location-dependent perceptions. Our neural architecture employs learnable origin embeddings that condition the interpretation of sequential geometric features, creating trajectory representations that preserve both commonalities and differences across starting locations.
Through contrastive learning objectives, the model learns to distinguish trajectories by origin while maintaining sensitivity to genuine spatial patterns. This decomposition enables systematic quantification of cognitive asymmetries through metrics such as origin divergence, measuring how distinct different starting locations feel cognitively, and global asymmetry, assessing how uniformly cognitive resources are distributed across space. Unlike traditional approaches that assume spatial equity, our framework reveals where urban design creates systematic cognitive inequalities and provides quantitative tools for measuring experiential fairness across neighborhoods.

\section{Methods}

\subsection{Problem Formulation}
We introduce origin-conditional trajectory encoding to quantify spatial perspective asymmetries in urban networks. Given $K$ origins, each origin $k$ associates with trajectory collection $\mathcal{T}_k = \{T_1^k, T_2^k, \ldots, T_{N_k}^k\}$ where trajectories $T_i^k = \{v_1, v_2, \ldots, v_L\}$ capture geometric features (visibility ratio, curvature). The challenge is learning representations that preserve origin-specific configurational signatures while identifying universal spatial patterns.The dataset is formulated as $\mathcal{D} = \{(\mathcal{T}_k, k)\}_{k=1}^K$.

\subsection{Conditional Trajectory Encoder}
Our architecture introduces origin-conditional processing through learnable origin embeddings that guide trajectory interpretation. Each trajectory undergoes bidirectional LSTM encoding to extract temporal patterns: where $\mathbf{H} = \text{BiLSTM}(T)$ and $\mathbf{H} \in \mathbb{R}^{L \times d_h}$
Origin $k$ maps to embedding $\mathbf{e}_k \in \mathbb{R}^{d_e}$ encoding configurational context. The key innovation is conditional attention that allows origin information to selectively weight trajectory features:
\begin{equation}
    \mathbf{c} = \text{Attention}(\mathbf{H}, \mathbf{e}_k) = \text{softmax}\left(\frac{\mathbf{H}\mathbf{e}_k^T}{\sqrt{d_e}}\right)\mathbf{H}
\end{equation}
This produces origin-conditioned trajectory representations that capture how the same movement patterns are interpreted differently from different spatial positions.

\subsection{Decomposed Representation Learning}
The conditioned features undergo fusion and decomposition to separate universal and origin-specific components: $\mathbf{z} = f_{\theta}([\mathbf{c}; \mathbf{e}_k])$ and $    \mathbf{z}^{\text{shared}} = \mathbf{z}[:d_s], \quad \mathbf{z}^{\text{specific}} = \mathbf{z}[d_s:]$
The shared component $\mathbf{z}^{\text{shared}}$ captures universal urban movement patterns, while $\mathbf{z}^{\text{specific}}$ encodes origin-dependent configurational signatures. This explicit decomposition enables systematic quantification of spatial perspective asymmetries.

\subsection{Joint Training Framework}
We optimize four complementary objectives. Reconstruction loss ensures information preservation:
\begin{equation}
    \mathcal{L}_{\text{recon}} = \mathbb{E}_{T,k}[|g_{\phi}(\mathbf{z}) - T|_2^2]
\end{equation}
Origin-contrastive loss promotes within-origin clustering while encouraging between-origin separation:
\begin{equation}
    \mathcal{L}_{\text{contrast}} = \mathbb{E}_{i,j}[\max(0, |\mathbf{z}_i^{\text{specific}} - \mathbf{z}_j^{\text{specific}}|_2 - m + \mathbb{I}[k_i = k_j])]
\end{equation}
Shared regularization enforces consistency across origins: $    \mathcal{L}_{\text{shared}} = \mathbb{E}[|\mathbf{z}^{\text{shared}} - \mathbb{E}[\mathbf{z}^{\text{shared}}]|_2^2]$
Orthogonality constraint encourages representational diversity between origins: $    \mathcal{L}_{\text{ortho}} = \sum_{p \neq q} (\boldsymbol{\mu}_p^{\text{specific}})^T \boldsymbol{\mu}_q^{\text{specific}}$
The joint optimization balances trajectory fidelity with configurational differentiation, enabling meaningful asymmetry quantification through learned embedding distances.

Regarding the evaluation, spatial perspective asymmetries are quantified through statistical analysis of origin-specific embeddings. Origin divergence measures configurational differentiation: $D_{\text{origin}} = \sum_{k=1}^K \text{Var}(\boldsymbol{\mu}_k^{\text{specific}})$. And global asymmetry captures overall spatial inequality: $    A_{\text{global}} = \frac{1}{K(K-1)} \sum_{i \neq j} |\boldsymbol{\mu}_i^{\text{specific}} - \boldsymbol{\mu}_j^{\text{specific}}|_2$
These metrics provide quantitative assessment of how urban morphology creates systematic configurational asymmetries across different spatial positions.

\section{Results}
\begin{figure}[htbp]
    \centering
    \includegraphics[width=1\textwidth]{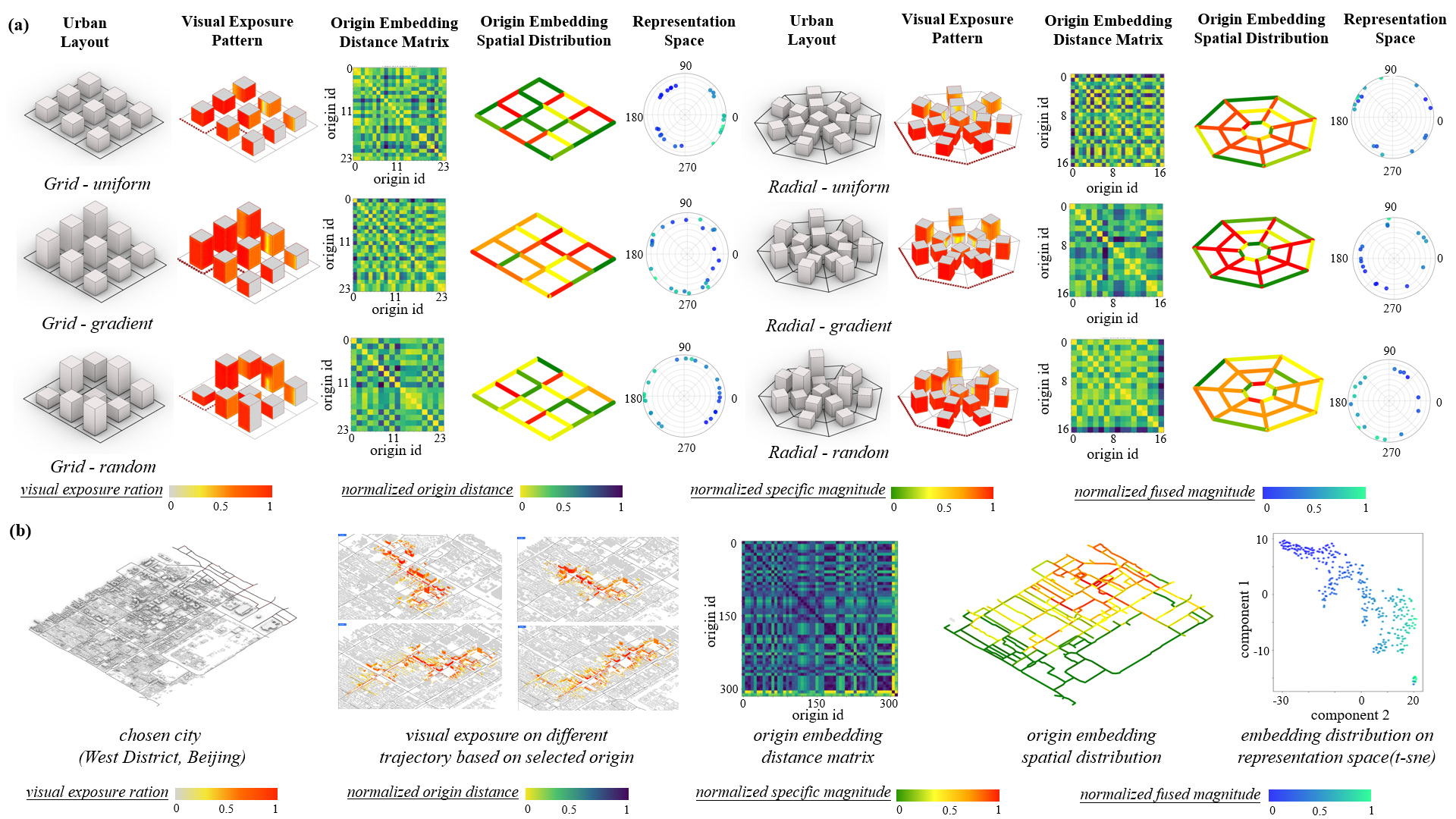}
    \caption{Learning and Visualization of Urban Spatial Representations based on Conditional Encoding}
    \label{fig:diagram1}
\end{figure}

We validate our conditional trajectory encoding framework across six synthetic cities and Beijing's Xicheng District (Fig.\ref{fig:diagram1}). The model trains with 80 epochs, batch size 16, and learning rate 0.002, achieving convergent reconstruction loss between 12-17\% across all urban configurations. Training stability is maintained through gradient clipping and early stopping.
\textbf{(1)Layout-Dependent Asymmetry Patterns: } Fig.\ref{fig:diagram1}a demonstrates systematic differences between grid and radial morphologies through five analysis dimensions. The Origin Embedding Distance Matrix reveals grid cities exhibit structured block patterns with high intra-block similarity, while radial cities show more distributed asymmetry patterns. Grid cities demonstrate 62\% variation in origin divergence across height distributions (1.32-2.14), with the distance matrices showing increasingly fragmented patterns from uniform to random heights. Radial cities maintain more consistent asymmetry levels (1.85-2.41, 30\% variation), with their distance matrices preserving radial structure regardless of height distribution.
\textbf{(2)Height Distribution Effects: } The Visual Exposure Patterns illustrate how building height variations create distinct configurational signatures. Grid-uniform shows regular, homogeneous exposure patterns corresponding to lowest asymmetry (origin divergence 1.32). Grid-gradient exhibits systematic directional gradients in the exposure map, producing moderate asymmetry (1.78). Grid-random demonstrates fragmented, heterogeneous exposure patterns resulting in maximum asymmetry (2.14). Radial cities maintain consistent radial exposure patterns across height variations, explaining their structural stability in asymmetry metrics.
\textbf{(3)Embedding Space Analysis: }The Origin Embedding Spatial Distribution  and Representation Space visualizations confirm successful origin-specific learning. Grid cities produce embedding networks that reflect their underlying orthogonal structure, with clear clustering patterns that vary by height distribution. The representation space distributions show grid-uniform creates compact clusters, while grid-random produces more dispersed patterns. Radial cities consistently generate circular embedding arrangements in both spatial distribution and representation space, indicating robust structural coherence across morphological variations.
\textbf{(4)Real-World Validation:} Beijing's Xicheng District (Fig.\ref{fig:diagram1}b) exhibits complex mixed-morphology characteristics across all analysis dimensions. The visual exposure maps show heterogeneous patterns combining regular grid-like areas with irregular historical sections. The origin embedding distance matrix reveals locally clustered asymmetries with high-contrast boundaries, consistent with transitions between different urban fabric types. The representation space analysis using t-SNE shows multi-modal clustering, confirming the mixed morphology hypothesis and validating the framework's applicability to complex real-world urban structures.

\section{Discussion}
Our empirical findings reveal a fundamental pattern overlooked in urban analytics: spatial positions within the same city create systematically different configurational experiences. The 62\% variation in origin divergence across grid positions and 30\% variation in radial cities demonstrate that these differences are not random fluctuations but structured asymmetries inherent to urban morphology. Current urban analysis methods treat such variations as noise to be averaged away, yet our results suggest they represent the essential mechanism through which urban structure operates. Grid cities achieve configurational equity only under uniform height conditions (origin divergence: 1.32), while height variations amplify positional inequalities, creating spatial positions that offer fundamentally different navigational and perceptual opportunities. This challenges the implicit assumption in urban modeling that spatial positions are functionally equivalent. Instead, our conditional encoding framework reveals that cities naturally organize themselves through configurational gradients—systematic differences in spatial information exposure that create the potential for movement, attention, and urban activity. Without such asymmetries, cities would lack the directional forces that guide flows and generate the patterns of use that constitute urban life. This perspective shift from symmetric to asymmetric urban analysis provides new theoretical foundations for understanding how spatial configuration drives urban dynamics.

\bibliography{references}
\end{document}